\documentclass[conference]{IEEEtran}
\usepackage{times}

\usepackage[numbers]{natbib}
\usepackage{multicol}
\usepackage[bookmarks=true]{hyperref}

\usepackage{url}
\usepackage{booktabs}       
\usepackage{graphicx}
\usepackage{multirow}
\usepackage{adjustbox}
\usepackage{wrapfig}
\usepackage{bbding}
\usepackage{xcolor}
\usepackage{amsmath}
\usepackage{amssymb}

\graphicspath{{figures/}}

\begin{document}
\title{3D Surface Reconstruction in the Wild by Deforming Shape Priors from Synthetic Data}

\author{\authorblockN{Nicolai Häni}
\authorblockA{University of Minnesota\\
Minneapolis, Minnesota 55455\\
Email: haeni001@umn.edu}
\and
\authorblockN{Jun-Jee Chao}
\authorblockA{University of Minnesota\\
Minneapolis, Minnesota 55455\\
Email: chao0107@umn.edu }
\and
\authorblockN{Volkan Isler}
\authorblockA{University of Minnesota\\
Minneapolis, Minnesota 55455\\
Email: isler@umn.edu}}

\maketitle

\begin{abstract}
Reconstructing the underlying 3D surface of an object from a single image is a challenging problem that has received extensive attention from the computer vision community. Many learning-based approaches tackle this problem by learning a 3D shape prior from either  ground truth 3D data or multi-view observations. To achieve state-of-the-art results, these methods assume that the objects are specified with respect to a fixed canonical coordinate frame, where instances of the same category are perfectly aligned. In this work, we present a new method for joint category-specific 3D reconstruction and object pose estimation from a single image. We show that one can leverage shape priors learned on purely synthetic 3D data together with a point cloud pose canonicalization method to achieve high quality 3D reconstruction in the wild. Given a single depth image at test time, we first transform this partial point cloud into a learned canonical frame. Then, we use a neural deformation field in to reconstruct the 3D surface of the object. Finally, we jointly optimize object pose and 3D shape to fit the partial depth observation. Our approach achieves state-of-the-art reconstruction performance across several real-world datasets, even when trained only on synthetic data. We further show that our method generalizes to different input modalities, from dense depth images to sparse and noisy LIDAR scans.
\end{abstract}
\IEEEpeerreviewmaketitle

\section{Introduction}
Surface reconstruction of a 3D object from a partial observation, such as a depth image or a LIDAR scan, is a longstanding problem in computer vision~\cite{yan_shapeformer_2022,yu_pointr_2021,mittal_autosdf_2022,sen_scarp_2023}. Discovering the full shape of an object from a partial input has many applications, including in visual servoing~\cite{kumar_pose_2017}, robotic manipulation~\cite{bylow_real-time_2013,mousavian_6-dof_2019,sen_scarp_2023}, autonomous driving~\cite{campbell_autonomous_2010,zeng_dsdnet_2020} and content creation~\cite{huang_3dlite_2017}.

Every computational approach aimed at 3D reconstruction must choose a representation for the D model.
An increasingly popular choice is to use neural fields~\citep{park_deepsdf_2019,mescheder_occupancy_2019} for this task. These neural fields, trained on 3D ground truth data, represent the de-facto gold standard regarding reconstruction quality. At inference time, the learned 3D shape prior is adapted to the partial observation. However, these methods suffer from two major limitations: i) they require 3D ground truth data in the form of occupancy values or signed distance functions and ii) these models expect shapes to be aligned and normalized in a fixed canonical coordinate frame - a frame of reference that is shared between all instances in the shape category. These two limitations have for now, limited these approaches to synthetic data, such as Shapenet~\cite{chang_shapenet_2015}. 

To remove the reliance on 3D data, the community has shifted to dense~\citep{mildenhall_nerf_2020}, or sparse~\citep{zhang_ners_2021} multi-view supervision with known camera poses, which can be estimated using Structure from Motion (SfM). Similarly, single-view 3D reconstruction methods have also made considerable progress by using neural fields as their shape representation~\citep{lin_sdf-srn_2020,duggal_topologically-aware_2022}. While these single-view methods can be trained from unconstrained image collections, they have not achieved the high quality of multi-view or 3D ground truth supervised models. In this work, we aim to answer the question: \textit{How can we achieve the reconstruction quality of 3D supervised methods from single view observations in the wild?}

\begin{figure}[ht!]
	\centering
	\includegraphics[width=\columnwidth]{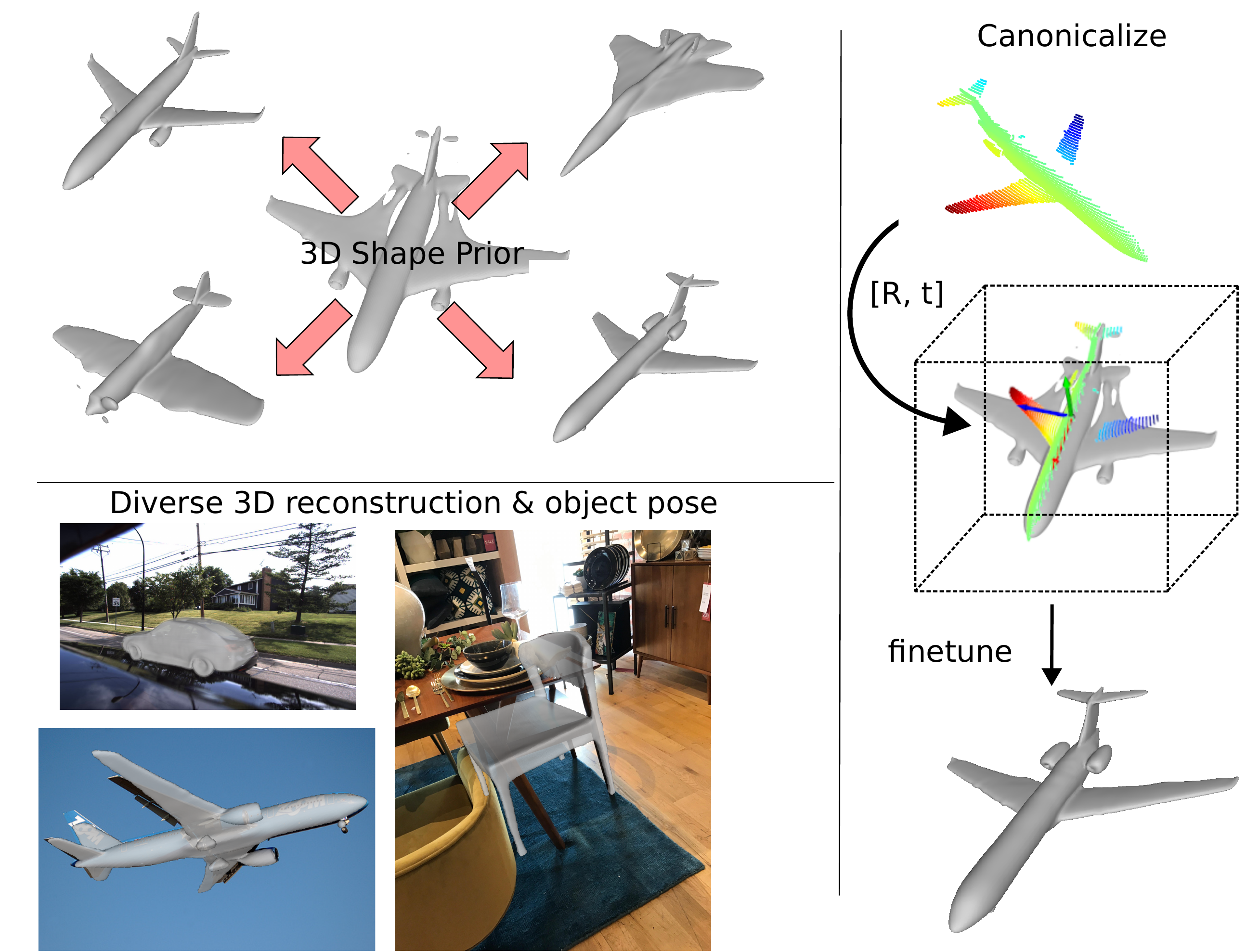}
	\caption{We pretrain a 3D shape prior on synthetic 3D data. Using the Equi-pose~\cite{li_leveraging_2021} canonicalization algorithm, we register the depth image to the canonical coordinate frame and use a finetuning scheme to jointly estimate the surface reconstruction and object pose from a single observation, leading to a model that generates diverse 3D reconstructions and object poses.}
	\label{fig:teaser}
\end{figure}

We propose to use a single depth image from a calibrated camera together with a pretrained canonicalization network to register the partial point clouds to the canonical coordinate space. We reduce the effect of errors in the canonicalization process by jointly fine-tuning the latent shape descriptor and object pose using only the partial observation as input (Figure~\ref{fig:teaser}). 
We achieve 3D reconstruction results on synthetic data close to or better than the state-of-the-art. Furthermore, we show that using depth images as input allows for generalization across various datasets, from dense depth in synthetic and natural images to sparse depth inputs from LIDAR scans. 

\section{Related Work}
3D object reconstruction based on a conditional input, such as images or depth is an active research area~\citep{chen_learning_2019, mescheder_occupancy_2019, mildenhall_nerf_2020, sitzmann_scene_2019, tulsiani_multi-view_2017, hani_continuous_2020}. The defacto gold standard in terms of reconstruction quality uses 3D ground truth data~\citep{park_deepsdf_2019,mescheder_occupancy_2019}. These approaches are largely limited to synthetic data, such as Shapenet~\citep{chang_shapenet_2015} as they require shapes that are aligned in a common canonical coordinate frame. Reconstruction of real-world shapes has been performed by transferring the learned representation across domains~\citep{duggal_mending_2022,bechtold_fostering_2021} or with the use of special depth sensors~\citep{newcombe_kinectfusion_2011, choe_volumefusion_2021}. However, collecting 3D ground truth data in the real world can be difficult. With the development of neural rendering and inverse graphics methods, the requirement for 3D ground truth has been relaxed in favor of dense multi-view supervision~\citep{xu_disn_2019,mildenhall_nerf_2020, goel_differentiable_2022, zhang_ners_2021} or single view methods that require ground truth camera poses for training~\citep{lin_sdf-srn_2020, duggal_topologically-aware_2022}. However, not all applications allow for the collection of multi-view images, and estimating camera poses from images remains challenging. With the advent of generative models for 3D shapes~\citep{gao_get3d_2022}, using 3D supervision has become an interesting prospect once more. Our work shows how we can leverage shape canonicalization for shape reconstruction in the wild.

\subsection{Pose Registration and 3D Shape Canonicalization}
Reliance on camera poses is an issue for many real-world datasets but a necessary step for neural rendering or deformation-based models. 
Point cloud registration can estimate the object pose directly and has achieved good performance when matching point clouds of the same object; however, these methods are unsuitable for single view pose estimation without a ground truth 3D model~\citep{jiang_sampling_2021,wu_feature_2021,qin_geometric_2022}. Category-level object pose estimation methods achieve tremendous results, for supervised training mechanisms~\citep{rempe_caspr_2020,novotny_c3dpo_2019, wang_normalized_2019,chao_category-level_2022}, and using only self-supervision ~\citep{spezialetti_learning_2020, sun_canonical_2021, li_leveraging_2021, sajnani_condor_2022, katzir_shape-pose_2022}. For example, Canonical Capsules~\citep{sun_canonical_2021} learn to represent object parts with pose-invariant capsules by training a  Siamese network in a self-supervised manner. Although the learned capsules can reconstruct the input point cloud in the learned canonical frame, Canonical Capsules only works on complete point clouds. In contrast, Equi-pose~\citep{li_leveraging_2021} can canonicalize both complete and partial point clouds. By leveraging an SE(3) equivariant network, Equi-pose simultaneously learns to estimate object pose and canonical point cloud completion. Our work shows that one can leverage Equi-pose with  test time pose refinement to get accurate shape reconstructions in canonical space.

\subsection{Pointcloud Completion}
Instead of relying on ground truth camera poses, we use depth images to register the partial 3D point cloud into a canonical frame. As we use depth images as input, our method closely relates to point cloud completion algorithms. Early work on point cloud completion used 3D convolutions to learn shape completion~\citep{dai_shape_2017,huang_pf-net_2020}. However, 3D convolutions are costly and operate on a canonical voxel grid. More recently, PointNet encoders were used for shape completion~\citep{liu_morphing_2020,yuan_pcn_2018}. Transformers have also been shown to work well on this task~\citep{yu_pointr_2021,yan_shapeformer_2022}. However, these methods rely on points already in a canonical coordinate frame. Further, these methods do not reconstruct the underlying surface of the object but output a limited number of points. In contrast, out method does not rely on canonical input points and reconstructs the underlying object surface with high fidelity.

\subsection{Surface Reconstruction from a Single View}
There have been extensive studies on 3D reconstruction from single view images using various 3D representations, such as voxels~\citep{yan_perspective_2016,tulsiani_multi-view_2017,wu_learning_2018,yang_learning_2018,wu_learning_2018,wu_marrnet_2017}, points~\citep{fan_point_2017,yang_pointflow_2019}, primitives~\citep{deng_cvxnet_2020, chen_bsp-net_2020} or meshes~\citep{kanazawa_learning_2018, goel_differentiable_2022}. Most of the methods above use explicit representations, which suffer from limited resolution or fixed topology. Neural rendering and neural fields provide an alternative representation to overcome these limitations. Recent methods showed how to learn Signed Distance Functions (SDFs)~\citep{xu_disn_2019,lin_sdf-srn_2020,duggal_topologically-aware_2022} or volumetric representations such as occupancy~\citep{ye_shelf-supervised_2021}, which have shown great promise in learning category-specific 3D reconstructions from unstructured image collections. However, these methods usually require additional information, such as ground truth camera poses or aligned 3D shapes, which limits their applicability. In our work, we propose a method that does not require ground truth camera poses or aligned 3D data and leverages widely available synthetic data to learn a category-specific 3D prior model.

\subsection{Learning Shape Reconstruction through Deformation}
Learning a generalizable model that maps a low-dimensional latent code to 3D surfaces can suffer from low-quality reconstructions. Category-specific deformable shape priors are useful to improve the quality of the  reconstruction~\citep{blanz_morphable_1999,engelmann_samp_2017,kanazawa_learning_2018,kar_category-specific_2015,loper_smpl_2015, mitchell_higher-order_2019}. These methods generally learn the deformation to an initial base shape. More recent work has used neural rendering together with SDFs~\citep{lin_sdf-srn_2020, duggal_topologically-aware_2022} to learn 3D shape priors from image collections and their associated camera poses. Other methods~\citep{deng_deformed_2021,zheng_deep_2021} jointly learn the deformation and the template shape in a canonical frame. In this work, we go one step further and show how we can leverage template shape and deformation models for incomplete observations registered to the template coordinate frame.

\section{Method}
Given a single segmented RGB-D image of an object, our goal is to jointly estimate the object pose and reconstruct the underlying 3D surface. To do so, we first learn a category-specific 3D template in the canonical coordinate frame together with an instance specific deformation field by leveraging synthetic 3D data. During test time, rather than directly reconstructing the shape in the camera coordinate frame, we use recent advances in point cloud canonicalization to transform a partial depth scan to the canonical space for surface reconstruction. Next, we describe first how we learn the 3D shape prior purely on synthetic data. Then we discuss how we reconstruct the surface of an observed depth image by deforming the learned canonical template shape.

\begin{figure*}[ht!]
	\centering
	\includegraphics[width=\textwidth]{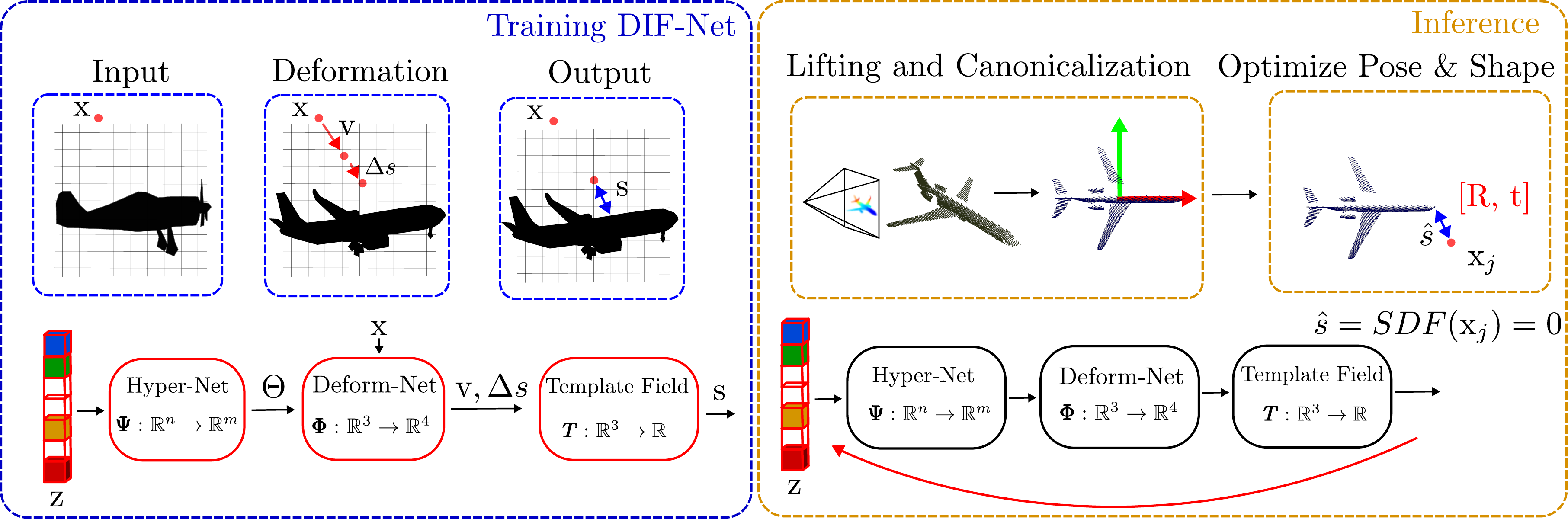}
	\caption{We train a DIF-Net as our 3D representation on purely synthetic data in an auto-decoder fashion. During inference, we lift the depth image onto 3D using the known camera intrinsics and estimate an initial transformation between the camera frame and the canonical frame. We jointly optimize the object pose and 3D shape to fit the partial observation. Trainable parameters/network parts are marked in red.}
	\label{fig:method_overview}
\end{figure*}

\subsection{3D Shape Prior}
Given a set of 3D objects ${\mathcal{O}_i}$, our goal is to learn a category-specific 3D shape prior together with a latent space describing the variation in shapes. Instead of directly mapping a low dimensional latent code $z_i \in \mathbb{R}^n$ to the 3D shape, we follow recent advances in learning 3D shape priors through deformation of a canonical template~\citep{zheng_deep_2021,deng_deformed_2021}. We jointly learn our 3D shape prior, represented as a neural network, and latent codes $z_i$ through the auto-decoder framework presented in~\cite{park_deepsdf_2019}. 
To generate high-quality 3D reconstructions, we use signed distance fields (SDFs). SDF is a function that assigns each point $x_j \in \mathbb{R}^3$ a scalar value $s_j \in \mathbb{R}$
\begin{equation}
SDF(x_j) = s_j, 
\end{equation}
 representing the distance to the closest object surface. The sign of $s_j$ indicates whether a point is inside (negative) or outside (positive) of the object, and the surface can be extracted as a mesh through marching cubes~\citep{lorensen_marching_1987}.  
We use a DIF-Net as our 3D shape prior network~\citep{deng_deformed_2021}.
The 3D representation network consists of a neural template field and a deformation field. We use the template field to capture common structures among a category of shapes by keeping the weights shared across all instances in the training set. The template field takes a 3D coordinate $x_j$ as input and predicts the signed distance to the closest surface $\hat{s}$:
\begin{equation}
T: x_j \in \mathbb{R}^3 \rightarrow \hat{s} \in \mathbb{R}
\end{equation}
To deform the template to a specific object instance, we use a deformation field together with a structural correction field
\begin{equation}
D: x_j \in \mathbb{R}^3 \rightarrow (v, \Delta s) \in \mathbb{R}^4.
\end{equation}
The vector $v$ deforms a point in the instance space to the template space, and the correction factor $\Delta s$ modifies the SDF value of point $x_j$ if it still differs from the ground truth value. The correction factor has been shown to be beneficial for categories with significant shape variations. For example, for chairs, there exist instances with and without armrests.
We use a Hyper-Network~\citep{sitzmann_scene_2019,mitchell_higher-order_2019} to condition the deformation field on a latent code. With a learned template field $T$ and deformation field $D$, the SDF value of a point $x_j$ can be obtained with 
\begin{equation}
s_j = T(x_j+v) + \Delta s = T(x_j + D_v(x_j)) + D_\Delta(x_j).
\end{equation}

\subsection{Training DIF}
During training we use the auto-decoder framework~\citep{sitzmann_scene_2019, park_deepsdf_2019} to jointly learn latent codes $z_i$ and the weights of the DIF network that predicts SDF values $\hat{s} = \Psi(x)$. Given a collection of shapes with ground truth SDF values on the object surface and in free space, we first apply an SDF regression loss from~\cite{sitzmann_implicit_2020} as
\begin{multline}
    \mathcal{L}_{sdf} =\sum_i (\sum_{x \in \Omega}|\Psi_i(x) - s| + \sum_{x \in S_i}(1-\langle\nabla\Psi_i(x), n\rangle) \\ + \sum_{x\in \Omega} || |\nabla\Psi_i(x)| - 1|| + \sum_{p\in\Omega  \setminus S_i}\rho(\Psi_i(x))),
    \label{equ:sdf}
\end{multline}

where $s$ and $n$ denote the ground truth SDF value and normal, $\nabla$ is the spatial gradient of the neural field, $\Omega$ is the 3D space in which values are sampled, and $s_j$ is the shape surface. We select an equal number of surface and free space points uniformly at random to compute this loss. The first term in Equ.~\ref{equ:sdf} regresses the SDF value; the second term learns consistent normals on the shape surface, the third term is the eikonal equation that enforces unit norm or the spatial gradients, and the last term penalizes SDF values close to $0$ which are far away from the object surface with $\rho(s) = \exp(-\delta \cdot |s|), \delta >>1$. For more details on this loss, check~\cite{sitzmann_scene_2019}. We further apply multiple regularization terms to help learn smooth deformations and consistent latent space. The first regularization term applies $L_2$ regularization on the embeddings as $\mathcal{L}_{z} = \sum_i||z_i||_2$.
Prior work by~\cite{deng_deformed_2021} showed that learning a template shape that captures common attributes across a category is improved by enforcing normal consistency across all shapes by regularizing the normals of the template networks with
\begin{equation}
    \mathcal{L}_{normal} = \sum_i \sum_{x\in S_i}(1 - \langle\nabla T(x+ D_v(x)), n\rangle).
\end{equation}
We further want deformations to be smooth and the optional corrections to the SDF field to be small, which is enforced with the following two loss terms $\mathcal{L}_{smooth} = \sum_i\sum_{x\in\Omega}||\nabla D_v(x)||_2$ and $\mathcal{L}_{c} = \sum_i \sum_{x \in \Omega} | D_{\Delta s}(x)|$. The overall loss to training the 3D shape prior is
\begin{equation}
    \mathcal{L} = \mathcal{L}_{sdf} + \lambda_1 {L}_{normal}  + \lambda_2 \mathcal{L}_{z} + \lambda_3 \mathcal{L}_{smooth}+ \lambda_3 \mathcal{L}_{c},
\end{equation}
with the $\lambda$ terms weighing the relative importance of each loss term.

\subsection{Point Cloud Lifting and Canonicalization during testing}
During inference, we use a single RGB-D image and known camera intrinsic parameters to lift the depth image to a partial 3D point cloud. In order to predict the deformation field, we first transform the partial point cloud to the canonical coordinate frame using Equi-pose~\citep{li_leveraging_2021} as our pose estimation module. 
Equi-pose is a SE(3)-equivariant network that learns category-specific canonical shape reconstruction and pose estimation in a self-supervised manner. By enforcing consistency between the invariant shape reconstruction and the input point cloud transformed by the estimated pose,  Equi-pose can estimate the pose of the input point cloud with respect to the learned canonical frame.
Therefore, we first input a complete template shape in our canonical frame to Equi-pose such that the transformation between our canonical frame and Equi-pose's canonical frame can be obtained. This way, we can transform any observed partial point cloud to our canonical frame using Equi-pose as a pose estimator. However, the estimated pose from Equi-pose can only serve as a noisy initialization. We show in Section~\ref{subsec:joint} how our method further finetunes the pose to achieve accurate shape reconstruction.

\subsection{Jointly Optimizing Shape and Pose}
\label{subsec:joint}
Once we train the 3D shape prior network and the partial input point cloud is roughly aligned in the canonical space, we reconstruct the object surface by optimizing the latent code and the object pose while keeping the SDF network weights fixed. As canonical 3D reconstruction methods are sensitive to minor deviations between estimated and canonical coordinate frames, we jointly optimize the latent code $z_i$ and the initial transformation by minimizing the SDF values at the observed depth points. At the same time, we sample random points in free space for the Eikonal term to ensure that the neural field is an SDF. We represent the translation as a three-dimensional vector initialized to zero and use the continuous 6D rotation parametrization from~\cite{zhou_continuity_2019} for rotations. We choose a random latent code from the learned latent space as our initialization $z_i$ and optimize

\begin{equation}
    \text{min}_{z_i, R, t} \mathcal{L}_{sdf} + \lambda_2 \mathcal{L}_z.
\end{equation}
\section{Experiments}
\textbf{Datasets} According with other works in the literature, we include three categories in our experiments: car, chair, and airplane. These categories are present across multiple datasets, facilitating comparison between approaches and enabling evaluation of a methods generalization capabilities. We use synthetic data from the ShapeNet dataset~\citep{chang_shapenet_2015} to train our deformation and template networks using 3D ground truth. Then our method trained on Shapenet is directly evaluated on the following datasets: ShapeNet~\citep{chang_shapenet_2015}, Pix3D chairs~\citep{lim_parsing_2013}, Pascal3D+~\citep{xiang_beyond_2014} and the DDAD~\citep{guizilini_3d_2020}, without retraining. Since Pascal3D+ and Pix3D do not contain depth scans, we generate the partial point clouds by removing invisible points of the CAD models using the ground truth camera poses. As DDAD does not provide reconstruction ground truth, we show the performance of our method on real-world noisy scans qualitatively only. See the appendix for additional information on datasets, baselines, and implementation details.

\textbf{Implementation Details} In line with prior work we train the 3D shape network on three categories in the Shapenet~\citep{chang_shapenet_2015} dataset, namely \textit{car, chair} and \textit{plane}. The networks are trained using the Adam optimizer~\citep{kingma_adam_2014}. We use batch size $128$ shapes per iteration and use $4000$ points on the surface, and $4000$ randomly sampled points in free space per object. Training takes $10$ hours on four NVIDIA V100 GPUs.

\textbf{Baselines} We compare against the state-of-the-art in single view, category-specific 3D object reconstruction: i) SDF-SRN~\citep{lin_sdf-srn_2020}, a neural field method that represents the object in camera coordinate frame and uses a neural renderer with silhouette supervision. ii) TARS-3D~\cite{duggal_topologically-aware_2022} is a method that uses ground truth camera poses to render a deformed template shape in canonical space to the image coordinate frame. Note that TARS-3D does not require ground truth camera poses during inference, but also does not estimate the object pose. Rather, it outputs the estimates surface reconstruction in the canonical coordinate frame. Instead, both TARS and SDF-SRN output the 3D surface reconstruction in the canonical coordinate frame. 

As our method is closely related to point cloud completion, we further compare our method against a transformer-based point cloud completion method, PoinTr~\citep{yu_pointr_2021}. In contrast to the baselines, our model is only trained on Shapenet and does not require camera pose information during both training and testing time. Only a partial depth scan is needed during inference.

\begin{figure*}[htbp!]
	\centering
	\includegraphics[width=\textwidth]{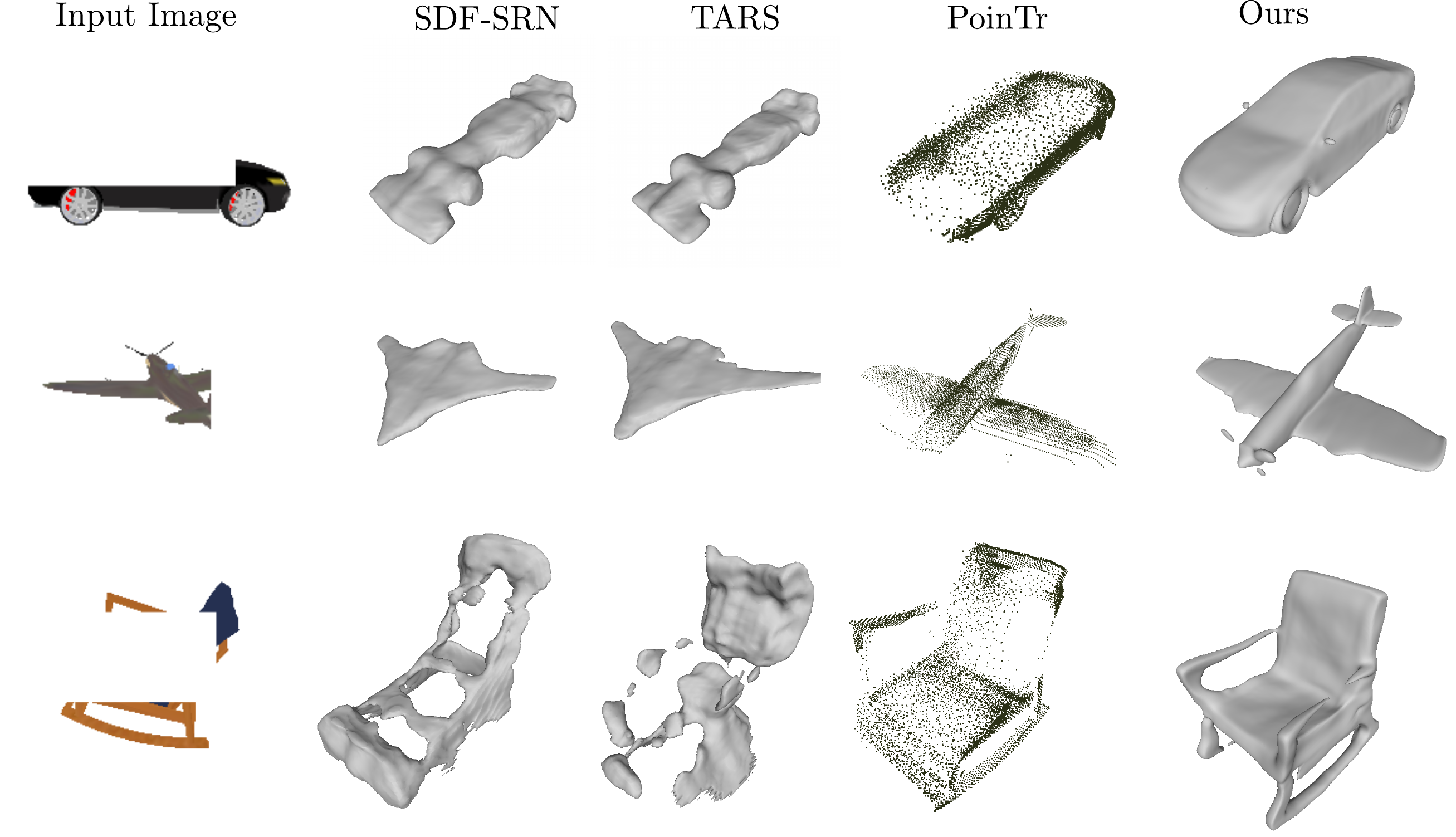}
	\caption{Qualitative result on the occluded Shapenet dataset. Our model outputs more accurate 3D shapes, as we do not condition on the input image, but can finetune the latent shape based on the partial depth observation.}
	\label{fig:occlusion}
\end{figure*}

\textbf{Evaluation Metrics} In this work, we follow~\cite{tatarchenko_what_2019} and report the F1-score at threshold $1\%$ as our primary evaluation metric. Tatarchenko et. al showed, that common metrics, such as chamfer distance and IoU allow for large variations from the ground truth model. In contrast, F1 score with a tight threshold requires the prediction to closely follow the ground truth to achieve high score. We additionally report the bidirectional chamfer distance (CD), multiplied by a factor of $1e4$ for readability.

\subsection{3D Reconstruction on synthetic Shapenet data}
Table~\ref{tab:results-shapenet} shows quantitative results of testing all approaches on the holdout test set of Shapenet. Our method outperforms the baselines in the car and plane categories with ground truth camera poses and is competitive in the chair category. We investigate cases where no ground truth camera poses are available and initialize our method and PoinTr with the Equi-pose estimates. Even without access to ground truth camera poses, our method performs comparable to or better than the baseline methods. We can see that PoinTr suffers greatly when the coordinates are not in the canonical coordinate system, showing that our approach of combining canonicalization with shape reconstruction is necessary for 3D shape reconstruction on real-world depth data. Our method's 3D reconstructions are more faithful to the underlying ground truth mesh, shown by the fact that we outperform all other methods on the F1 metric. PoinTr outputs only a limited number of points and does not reconstruct the underlying surface, nor does it give us correspondences between shapes in a category.

\begin{table}[htbp!]
	\caption{3D reconstruction results on synthetic test data. We report chamfer distance (CD) $\downarrow$ and F-score at threshold $0.01$ (F@$1\%$)$\uparrow$. $^\dagger$ with ground truth camera poses.}
	\label{tab:results-shapenet}
	\adjustbox{max width=\columnwidth}{
	\centering
		\begin{tabular}{l c c c c c c }
			\toprule
			Methods & \multicolumn{2}{c}{Car} & \multicolumn{2}{c}{Chair}  & \multicolumn{2}{c}{Plane} \\
			\cmidrule(lr){2-3} \cmidrule(lr){4-5} \cmidrule(lr){6-7}
			& CD $(\downarrow)$ & F@1 $(\uparrow)$ & CD $(\downarrow)$ & F@1 $(\uparrow)$ & CD $(\downarrow)$ & F@1 $(\uparrow)$\\ \midrule
			SDF-SRN   & 9.965 & 0.404 & 27.562 & 0.283 & 12.374 & 0.459 \\
		    TARS-3D & 10.175 & 0.412 & \textbf{28.823} & 0.272 & 11.302 & 0.418 \\
			PoinTr & 56.435 & 0.125 & 36.714 & 0.230 & 23.713 & 0.302 \\
			Ours & \textbf{9.371} & \textbf{0.439} & 32.138 & \textbf{0.334} & \textbf{16.562} & \textbf{0.631} \\
			\hline
			PoinTr $^\dagger$  & 13.249 & 0.264 & \textbf{12.834} & \textbf{0.352} & \textbf{3.637} & 0.685 \\
			Ours $^\dagger$  & \textbf{6.181} & \textbf{0.497} & 27.292 &0.343 & 4.495 & \textbf{0.768} \\
		\end{tabular}}
\end{table}

\begin{figure*}[ht!]
	\centering
	\includegraphics[width=\textwidth]{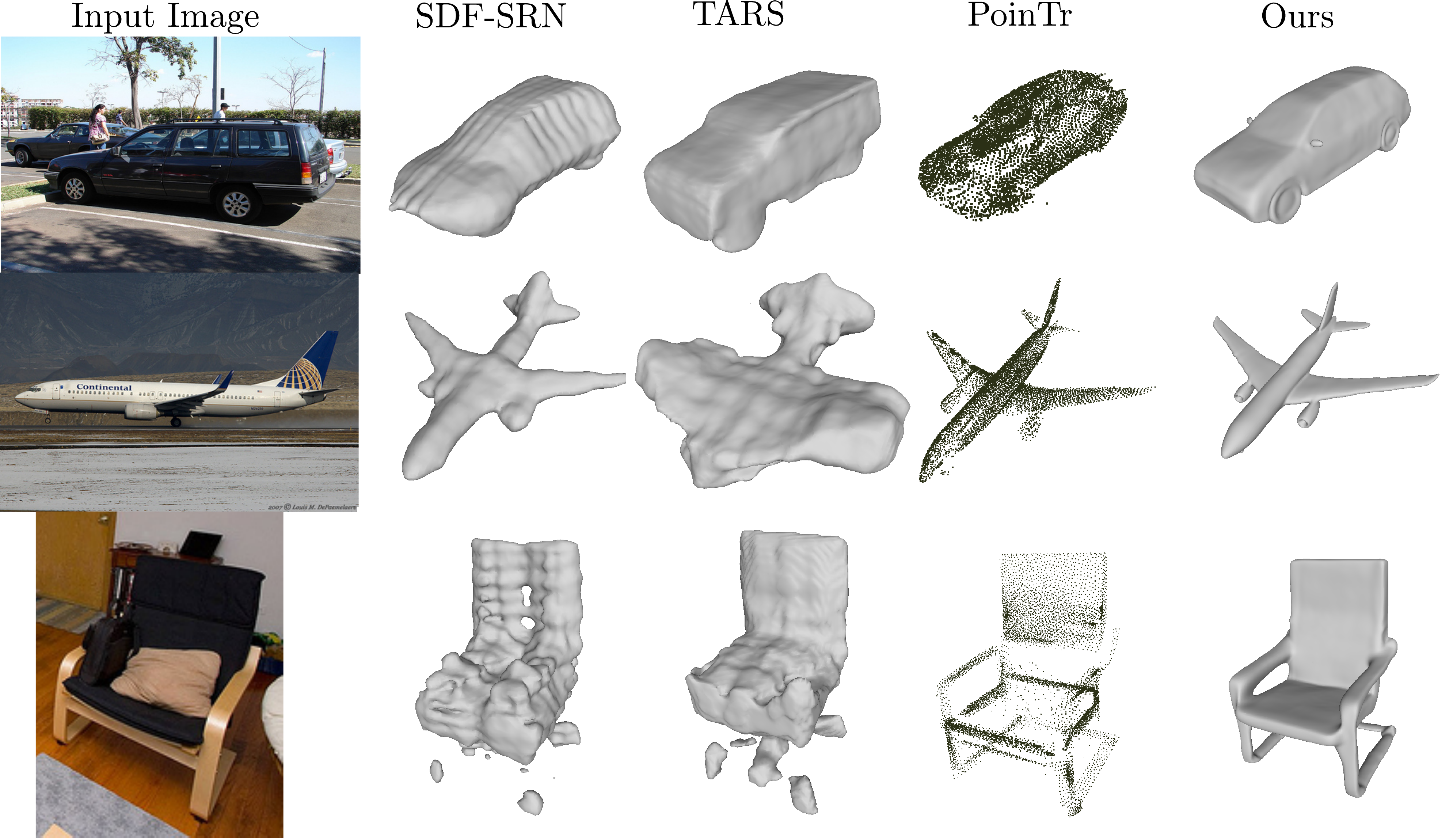}
	\caption{Qualitative results on the Pascal3D+ (top) and Pix3D (bottom) datasets.}
	\label{fig:qualitative_pascal}
\end{figure*}

\subsection{3D reconstruction on Shapenet with Occlusion}
This section investigates how our method performs when the observed object is occluded. For this experiment, we generate occluded area on the RGB-D images and remove the occluded points from the Shapenet test data as presented in Figure~\ref{fig:occlusion}. To study the influence of a wide variety of occlusions, we use rectangular overlap regions and generate overlap ratios from 5 - 85\%.  We apply models trained on Shapenet on the occlusion task for all methods. We remove the occluded region from the RGB-D images and canonicalize the occluded, partial pointclouds, before finetuning the surface reconstruction to fit the available points. As shown in Table~\ref{tab:results-occlusion}, our method outperforms all the baselines without access to ground truth camera poses despite not seeing any occluded data during training. Compared to the other baselines, PoinTr has a higher tolerance to occlusion. However, point cloud completion methods cannot predict the object's surface.
Moreover, these methods usually predict the complete point cloud by adding predicted points to the input. Therefore, the predicted point clouds are not guaranteed to be uniformly distributed and can have a higher density around the input points, resulting in lower chamfer distance, as shown in Figure~\ref{fig:occlusion}. In contrast, our method does not have these drawbacks and can reconstruct the occluded surface by leveraging the learned category-level prior with high fidelity in terms of F-score. 

\begin{table}[htbp!]
	\caption{3D reconstruction results on occluded data from the synthetic test set. We report chamfer distance (CD) $\downarrow$ and F-score at threshold $0.01$ (F@$1\%$)$\uparrow$. $^\dagger$ with ground truth camera poses.}
	\label{tab:results-occlusion}
	\adjustbox{max width=\columnwidth}{
    \centering
        \begin{tabular}{l c c c c c c }
			\toprule
			Methods & \multicolumn{2}{c}{Car} & \multicolumn{2}{c}{Chair}  & \multicolumn{2}{c}{Plane} \\
			\cmidrule(lr){2-3} \cmidrule(lr){4-5} \cmidrule(lr){6-7}
			& CD $(\downarrow)$ & F@1 $(\uparrow)$ & CD $(\downarrow)$ & F@1 $(\uparrow)$ & CD $(\downarrow)$ & F@1 $(\uparrow)$\\ \midrule
			\hline 
			SDF-SRN & 25.863  &  0.243 &  156.825 &  0.122 & 72.255 &  0.256 \\
			TARS-3D  & 23.806  & 0.241 & 108.976 & 0.131 & 58.998 &  0.263 \\
			PoinTr & 59.292 & 0.1225 & 54.327 & 0.184  & 37.425 & 0.322  \\
			Ours & \textbf{11.155} & \textbf{0.390} & \textbf{45.886} & \textbf{0.273}  & \textbf{22.223} & \textbf{0.580}\\
			\hline
			PoinTr $^\dagger$ & 22.110 & 0.208 & \textbf{19.079}  & \textbf{0.297} & \textbf{6.168} & 0.601 \\
			Ours$^\dagger$  &  \textbf{6.457} & \textbf{0.497} & {35.972} & \textbf{0.297} & {15.725} & \textbf{0.654} \\
		\end{tabular}}
\end{table}

\subsection{3D reconstruction on Pascal3D+ and Pix3D}
For this experiment, we test the generalization capabilities of our approach. We directly apply our model trained on Shapenet for reconstructing Pascal3D+ objects. TARS-3D and PoinTr also apply network weights trained on Shapenet to this task, while SDF-SRN is trained directly on the Pascal3D+ dataset.
As shown in Table~\ref{tab:results-pascal+pix3d}, our method again outperforms the other surface reconstruction methods without access to ground truth camera poses.  Figure~\ref{fig:qualitative_pascal} shows that our method generates reasonable outputs.

\begin{table}[htbp!]
	\caption{3D reconstruction results on the Pascal3D+ and Pix3D dataset. We report chamfer distance (CD) $\downarrow$ and F-score at threshold $1\%$ (F@$1\%$)$\uparrow$. $^\dagger$ with ground truth camera poses.}
	\label{tab:results-pascal+pix3d}
	\adjustbox{max width=\columnwidth}{
	\centering
		\begin{tabular}{l c c c c c c  c c }
			\toprule
            & \multicolumn{6}{c}{Pascal3D+} & \multicolumn{2}{c}{Pix3D}\\
			Methods& \multicolumn{2}{c}{Car} & \multicolumn{2}{c}{Chair}  & \multicolumn{2}{c}{Plane} & \multicolumn{2}{c}{Chair} \\
			\cmidrule(lr){2-3} \cmidrule(lr){4-5} \cmidrule(lr){6-7} \cmidrule(lr){8-9}
			& CD $(\downarrow)$ & F@1 $(\uparrow)$ & CD $(\downarrow)$ & F@1 $(\uparrow)$ & CD $(\downarrow)$ & F@1 $(\uparrow)$ & CD $(\downarrow)$ & F@1 $(\uparrow)$\\ \midrule
			SDF-SRN & 16.740 & 0.245 & 24.374 & 0.216 & 29.7457 & 0.169 & 60.432  & 0.158 \\
			TARS-3D& 19.129 & 0.2427 & 79.8922 & 0.1577 & 83.866 & 0.140 & 55.555 & 0.197\\
			PoinTr & 80.896 & 0.064 & 21.251 & 0.216 & \textbf{35.095} & 0.231 & \textbf{35.527} & 0.283\\
			Ours & \textbf{15.516} & \textbf{0.283} & \textbf{17.328} & \textbf{0.275} & 54.849 & \textbf{0.276} & 35.729 & \textbf{0.335}\\
			\hline
			PoinTr$^\dagger$ & 17.859 & 0.221 & \textbf{11.769} & \textbf{0.302} & \textbf{4.701} & \textbf{0.525} & \textbf{13.092} & \textbf{0.377}\\
			Ours$^\dagger$ & \textbf{13.370} & \textbf{0.302} & 16.804 & 0.274 & 54.286 & 0.289 & 31.165 & 0.335\\
		\end{tabular}}
\end{table}

We further test our method on another chair dataset, namely the chair category of the Pix3D dataset. As shown in Table~\ref{tab:results-pascal+pix3d}, our method outperforms the baselines. PoinTr again achieves a lower chamfer distance, which does not fully represent the reconstruction quality. As shown in Figure~\ref{fig:qualitative_pascal}, PoinTr generates point clouds that are not uniformly distributed while our method predicts smooth surfaces. 

\begin{figure*}[htbp!]
	\centering
	\includegraphics[width=\textwidth]{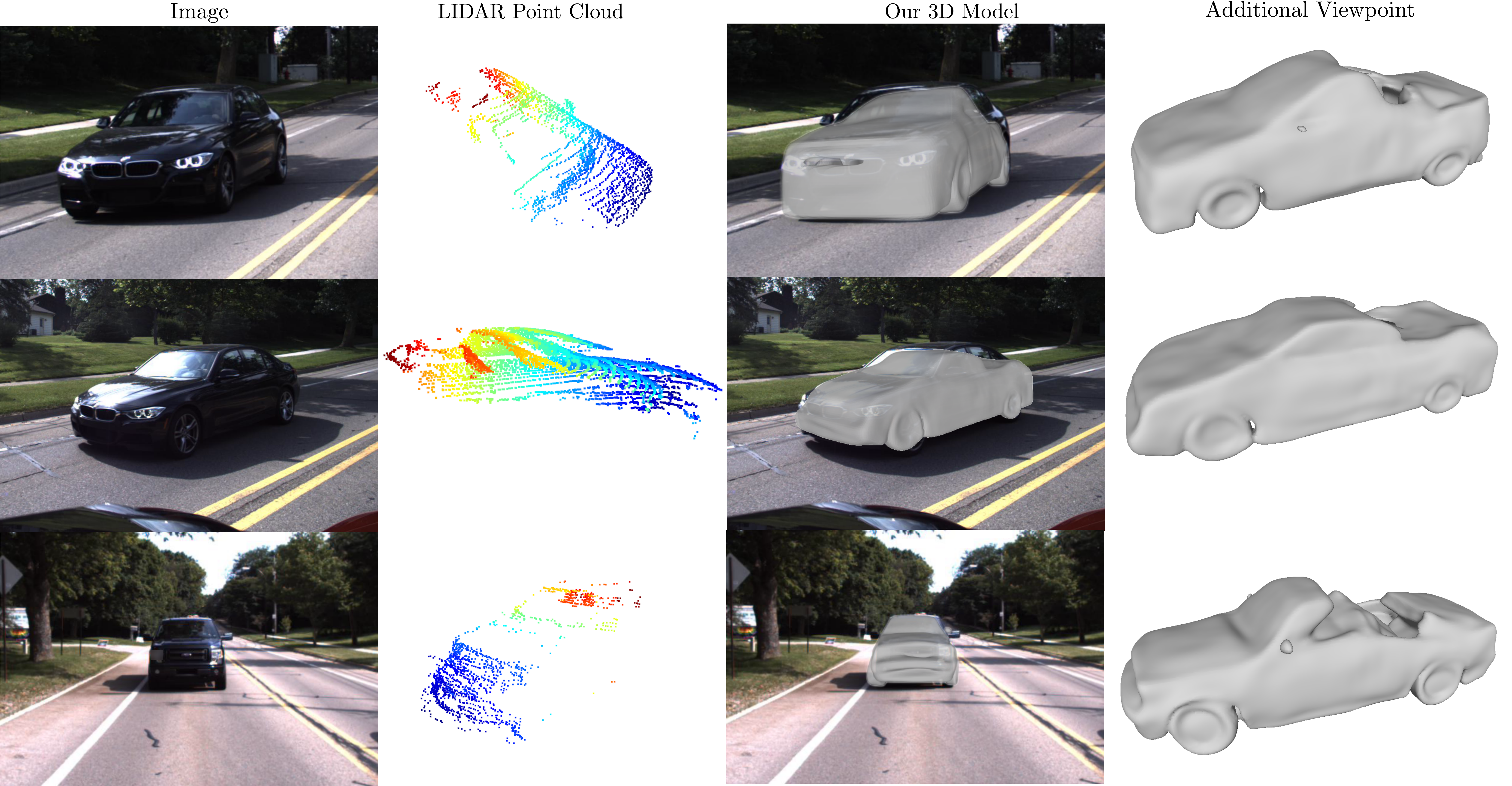} 
	\caption{Qualitative results on the DDAD dataset.}
	\label{fig:qualitative_ddad}
\end{figure*}

\subsection{Ablation Study and Failure Cases}
\label{sec:ablation}
In this section, we conduct an ablation study to asses the importance of optimizing the pose during inference. The main results are shown in Table~\ref{tab:ablation}. On the left, we show the ground truth mesh overlaid with the partial input points with pose optimization (blue) and without (green). We can see a noticeable reduction in F1 scores and a significant reduction in the reconstruction quality. The chamfer score difference between reconstruction methods across the dataset is slight, though, confirming~\cite{tatarchenko_what_2019} in that chamfer distance is not an ideal metric for 3D reconstruction. This ablation study shows that canonical reconstruction methods are sensitive to deviations in the estimated canonical coordinate frame. This result is also confirmed by the poor performance of PoinTr on input points where the estimated canonical coordinate frame is off (see Tables~\ref{tab:results-shapenet} and ~\ref{tab:results-pascal+pix3d}).

\begin{table}[ht!]
	\caption{Ablation of our method with and without pose optimization during inference. Left, we show the ground truth mesh overlayed with the optimized and non-optimized pose. We show the resulting optimized surfaces with and without pose optimization in the middle and right.}
	\label{tab:ablation}
	\centering
		\adjustbox{max width=\columnwidth}{
		\begin{tabular}{l | c | c }
		\toprule
		Pose optim. &  &  \checkmark \\
		\midrule
		CD & 23.008 & 22.223 \\
		\midrule
		F1  & 0.313  & 0.580 \\
		\midrule
		Shape \raisebox{-\totalheight}{\includegraphics[width=0.2\textwidth]{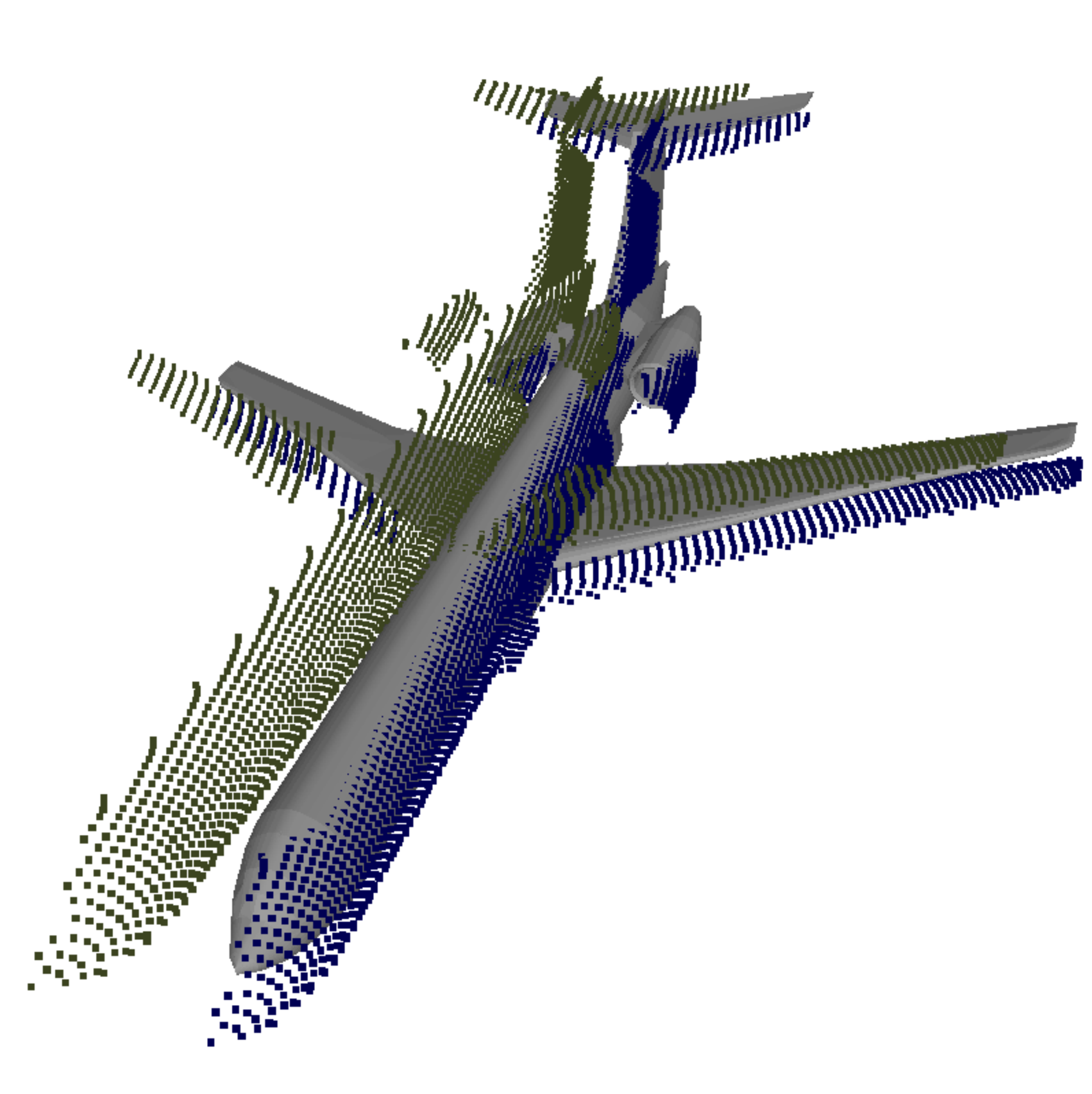}}& \raisebox{-\totalheight}{\includegraphics[width=0.2\textwidth]{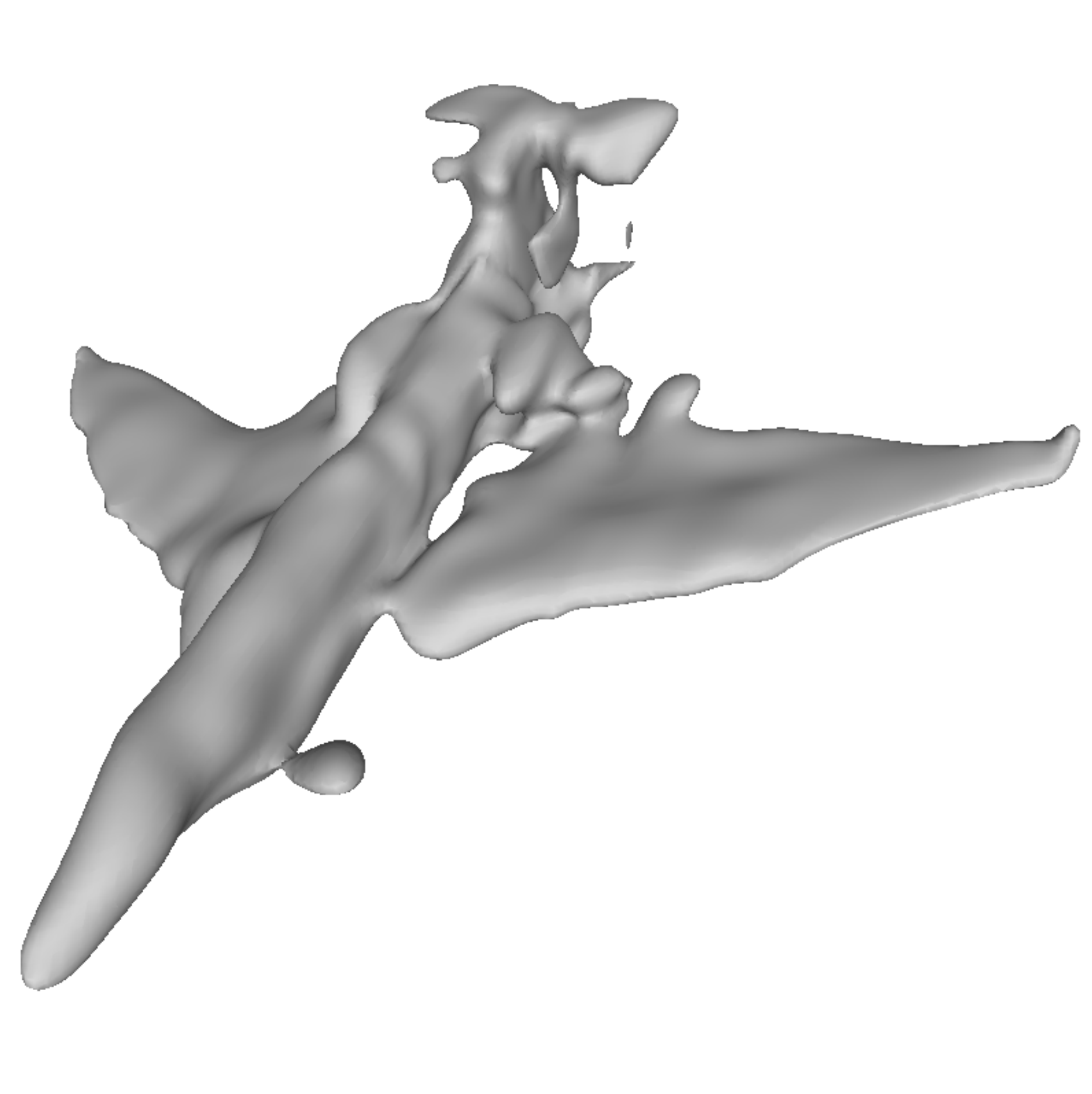}} & \raisebox{-\totalheight}{\includegraphics[width=0.2\textwidth]{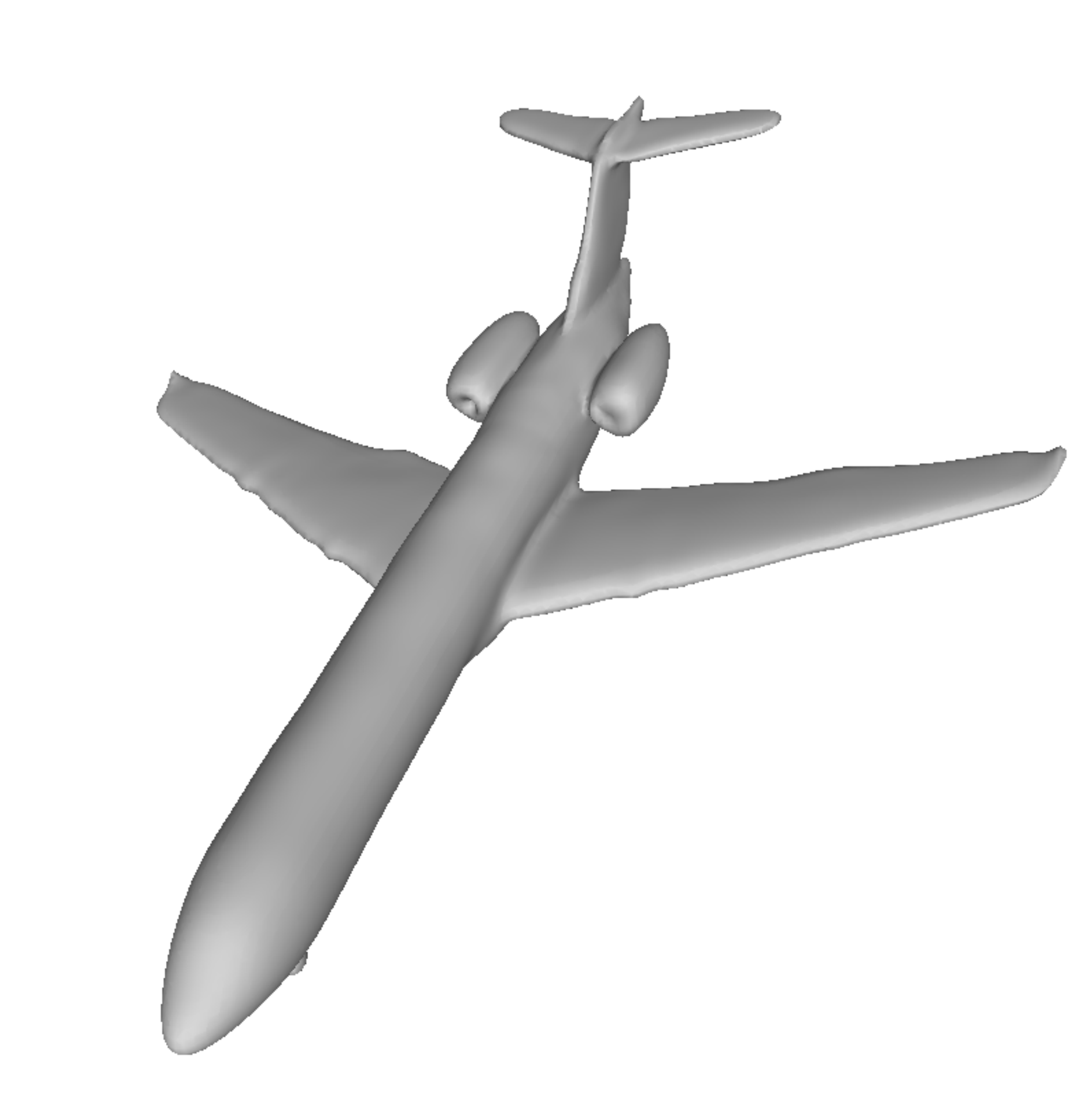}} \\ \\
		\bottomrule
		\end{tabular}}
\end{table}

\subsection{3D reconstruction on real-world noisy scans}
Finally, we apply our method trained on Shapenet directly to real-world noisy LIDAR scans. To demonstrate our tolerance to noise in the point clouds, we test our method on an autonomous driving benchmark DDAD~\citep{guizilini_3d_2020}. DDAD contains urban scenes scanned using LiDARs mounted on self-driving cars. To showcase our method, we extract frames that include other driving cars and crop the LiDAR scans of other cars with masked images. Finally, these noisy LiDAR scans are fed to the pose estimation module and our deformation field to reconstruct the surfaces. Since DDAD does not contain ground truth CAD models, we present the qualitative results in Figure~\ref{fig:qualitative_ddad}. Note that our method does not have access to the image but only the noisy LiDAR point clouds. Despite large portions of missing parts and the noise in the LiDAR scans, our method can still reconstruct reasonable car surfaces without access to ground truth camera poses.   
\section{Conclusion and future work}
We introduced a new method for complete 3D surface reconstruction of an object from real-world depth images. Our method relies on a representation obtained solely by training on synthetic data, which allows for extracting high-quality, category-specific geometry. We showed that even small errors in pose estimation lead to significant errors in 3D reconstruction. Therefore a simple method which uses an independently trained pose estimator followed by reconstruction in the object frame does not yield good reconstruction results. Instead, we presented a finetuning scheme to optimize the object surface and pose jointly during inference. We also showed that learning strong 3D priors benefits the 3D reconstruction of occluded objects. Our method generalizes across datasets and input modalities, from dense depth images to sparse LIDAR point clouds. While our process still exhibits failure modes when the error in the estimated pose is large, this could be alleviated by combining pose estimation and 3D reconstruction in an end-to-end trainable manner. We hope our work will inspire further work in this direction. 

\bibliographystyle{plainnat}
\bibliography{iclr2023_conference}

\appendix
\subsection{Dataset Details}
\textbf{Shapenet}~\cite{chang_shapenet_2015}:
For training on Shapenet, we follow~\cite{deng_deformed_2021} and extract 500,000 SDF values on the object's surface and 500,000 randomly sampled in a cube of side length 2. We also render a single RGB-D image with a camera sampled on the viewing hemisphere for testing according to~\cite{duggal_topologically-aware_2022}. We use the same train/test split for each category as DIF-Net~\citep{deng_deformed_2021}, but only keep the intersection of objects in both ours and the baselines test sets.

\textbf{Pascal3D+}~\cite{xiang_beyond_2014}:
Pascal3D+ is a real-world dataset containing camera images. The dataset provides object silhouettes, camera poses, and the CAD models used to annotate the camera poses. Since the same set of CAD models is used to annotate both the training and testing set, the dataset is considered to possess a bias~\citep{tulsiani_multi-view_2017}. However, unlike the other baselines, our method is only trained on Shapenet and therefore has not seen any of the CAD models from Pascal3D+ during training. We evaluate our method on the car, chair, and airplane category of Pascal3D+. We follow the testing split that is used by SDF-SRN~\citep{lin_sdf-srn_2020}. To generate partial point clouds in the camera frame as the input for our method, we first transform the CAD models into the camera frame using the ground truth camera poses. We remove points that are invisible from the camera origin. We do not have access to either ground truth poses or the complete CAD model during training and testing. 

\textbf{Pix3D}~\cite{lim_parsing_2013}:
Like Pascal3D+, Pix3D contains real-world 2D images annotated with 3D CAD models. Unlike Pascal3D+, Pix3D uses a variety of CAD models that align better with the images. We randomly select 200 images from the Pix3D chair dataset as the test set. We follow the same process as Pascal3D+ to generate the partial point clouds in the camera frame. Again, our method does not access the complete CAD models, and ground truth camera poses.

\textbf{DDAD}~\cite{guizilini_3d_2020}:
DDAD is an autonomous driving benchmark containing diverse urban scenes captured using car sensors. The dataset includes RGB videos captured using six cameras covering the 360-degree surrounding of the vehicle. DDAD also provides depth data across an entire 360-degree field of view scanned using long-range LiDAR sensors. We test our method on one scene with 100 frames. We extract only the frames that contain other cars in the camera's field of view. To generate the partial point clouds of the observed cars, we crop the LiDAR data with the ground truth poses and the masked images. The cropped LiDAR scans in the camera frame serve as our method's input. The DDAD dataset does not provide ground truth models of the observed cars, and our method does not have access to the ground truth camera poses. We therefore show only qualitative results. 

\subsection{Implementation Details}
\textbf{DIF-Net Architecture} DIF-Net uses a SIREN~\citep{sitzmann_implicit_2020} network as the MLP backbone for both, the deformation and template networks. The Hyper-network is a ReLU network, where each MLP predicts the weights of one of the layers in the deformation network $D$. 

\textbf{Training Details}
We initialize the latent codes to small values from $\mathcal{N}(0, 0.01)$. Each model is optimized for 60 epochs and during each epoch, the model has access to 200,000 free and 200,000 surface points. We set the weights for the SDF loss $\mathcal{L}_{sdf}$ to 3e3, 1e2, 5e1 and 5e2 according to~\cite{sitzmann_implicit_2020}. We follow~\cite{deng_deformed_2021} in choosing the weighting parameters as follows. $\lambda_1$ and $\lambda_4$ are 1e2 and 1e6 for all categories. $lambda_2$ is 5, 2, 5 for \textit{car, plane and chair}. $\lambda_3$ is 1e2, 1e2 and 5e1 for each of the above categories.

\textbf{Inference Details}
We jointly optimize the object pose and shape using the Adam optmizer with a learning rate of $0.001$ for the shape and $0.01$ for the pose. We optimize each shape for a total of 30 iterations, taking roughly 4 seconds.

\textbf{Equi-pose}~\cite{li_leveraging_2021}:
In this paper, we directly apply Equi-pose as an off-the-shelf pose estimation module. Since our method does not require accurate camera poses but rough initialization, we use the model weights trained on ModelNet40~\citep{wu_3d_2015} provided by the authors for all the experiments in this paper.

\subsection{Baselines}
\textbf{SDF-SRN}~\cite{lin_sdf-srn_2020}:
We directly use the source code and pre-trained weights for Pascal3D+ and Shapenet provided by the original authors. In this paper, we follow SDF-SRN's test split for Pascal3D+ dataset. As for Shapenet, we take the union between our test set and SDF-SRN's test set as the Shapenet test set. 

\textbf{TARS-3D}~\cite{duggal_topologically-aware_2022}:
We use the original implementation of TARS-3D from the authors. Since TARS-3D follows the same dataset setup as SDF-SRN, we use the network weights trained on Shapenet provided by the authors. 

\textbf{PoinTr}~\cite{yu_pointr_2021}:
We directly adopt the original code and network weights trained on Shapenet provided by the authors. For fair comparison, we transform the input partial point clouds into their coordinate frame with both ground truth and estimated camera poses. Furthermore, we follow their original training setup where the input point clouds are downsampled to 2048 points using farthest point sampling, and predict 8192 points as the complete point cloud.

\subsection{Additional Results}
\textbf{Extracted Template Shapes}
Here we show the extracted template shapes from our pretrained 3D prior network. We extract the zero-level set of the neural field using marching cubes. We can observe that the template represents shapes close to the "mean" object for cars and chairs. For planes however, the neural field does not represent an object, but fuses common aspects of different shapes together. 

\begin{figure}[ht!]
	\centering
	\includegraphics[width=\linewidth]{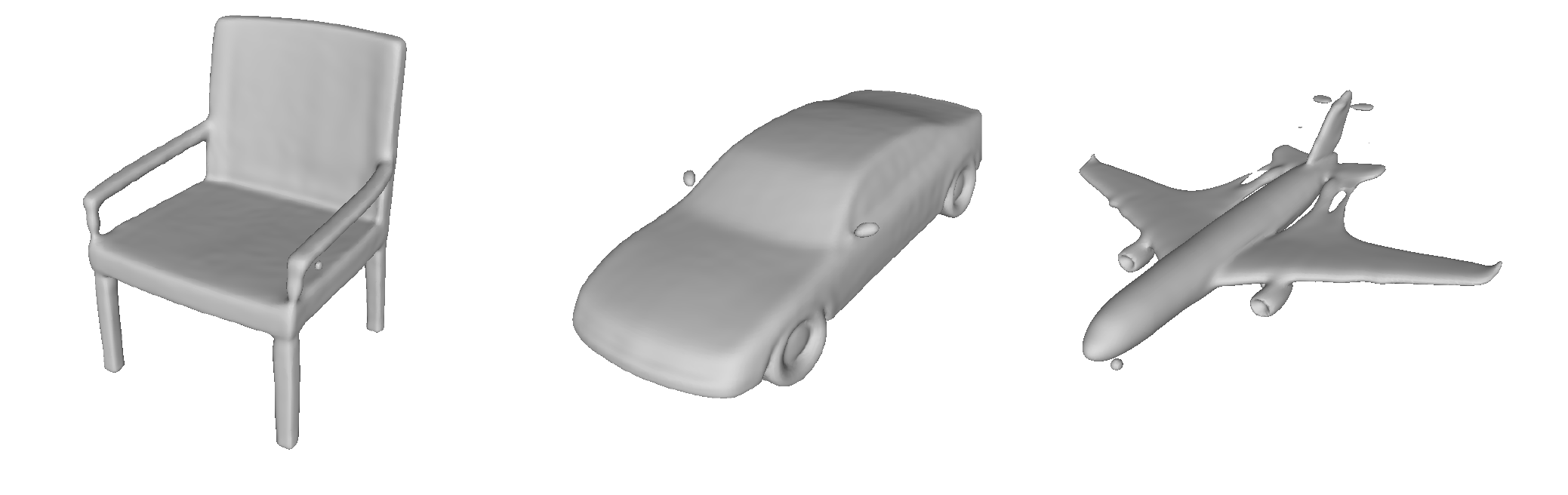}
	\caption{Extracted learned template shapes}
	\label{fig:template}
\end{figure}

\begin{figure*}[ht!]
	\centering
	\includegraphics[width=\textwidth]{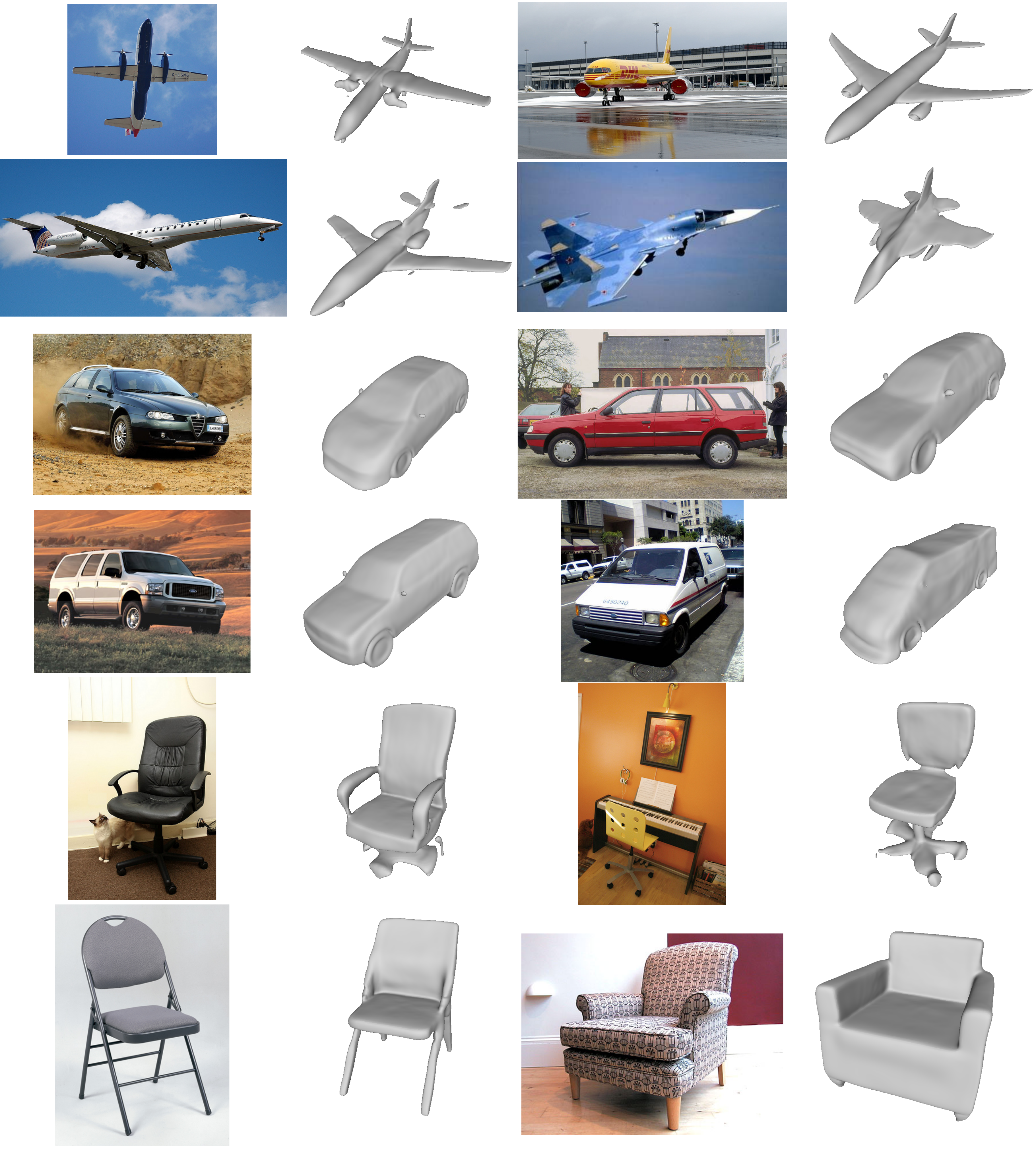}
	\caption{Additional results on the Pascal3D+ dataset}
\end{figure*}

\begin{figure*}[ht!]
	\centering
	\includegraphics[width=\textwidth]{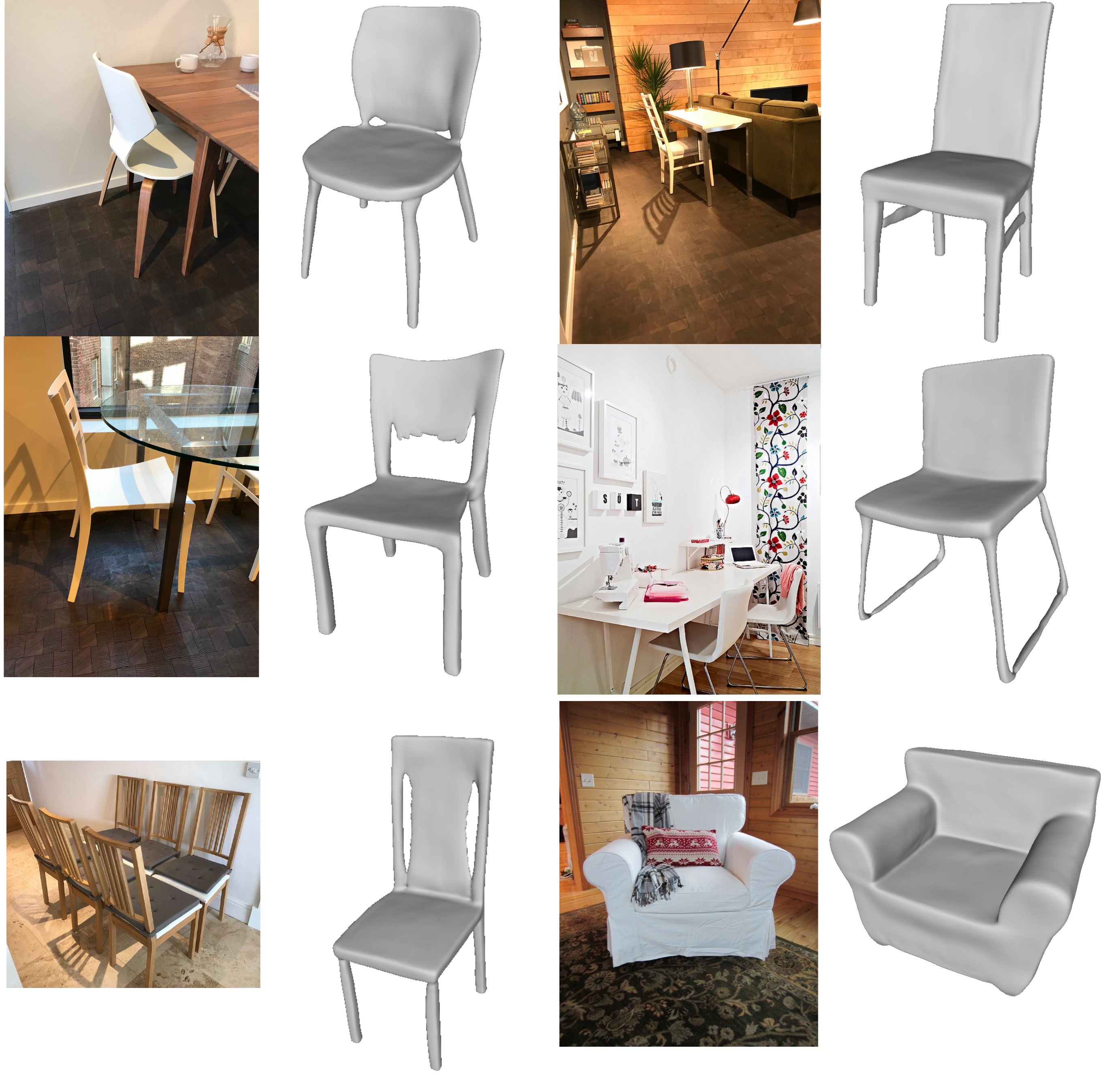}
	\caption{Additional results on the Pix3D dataset}
\end{figure*}

\end{document}